# Effects of Design and Hydrodynamic Parameters on Optimized Swimming for Simulated, Fish-inspired Robots

Donghao Li[1], Hankun Deng[1], Yagiz E. Bayiz[1], Bo Cheng[1], *Member, IEEE*

*Abstract*—In this work we developed a mathematical model and a simulation platform for a fish-inspired robotic template, namely Magnetic, Modular, Undulatory Robotics (μBots). Through this platform, we systematically explored the effects of design and fluid parameters on the swimming performance via reinforcement learning. The mathematical model was composed of two interacting subsystems, the robot dynamics and the hydrodynamics, and the hydrodynamic model consisted of reactive components (added-mass and pressure forces) and resistive components (drag and friction forces), which were then nondimensionalized for deriving key "control parameters" of robot-fluid interaction. The μBot was actuated via magnetic actuators controlled with harmonic voltage signals, which were optimized via EM-based Policy Hyper Parameter Exploration (EPHE) to maximize swimming speed. By varying the control parameters, total 36 cases with different robot template variations (Number of Actuation (NoA) and stiffness) and hydrodynamic parameters were simulated and optimized via EPHE. Results showed that wavelength of optimized gaits (i.e., traveling wave along body) was independent of template variations and hydrodynamic parameters. Higher NoA yielded higher speed but lower speed per body length however with diminishing gain and lower speed per body length. Body and caudal-fin gait dynamics were dominated by the interaction among fluid added-mass, spring, and actuation torque, with negligible contribution from fluid resistive drag. In contrast, thrust generation was dominated by pressure force acting on caudal fin, as steady swimming resulted from a balance between resistive force and pressure force, with minor contributions from added-mass and body drag forces. Therefore, added-mass force only indirectly affected the thrust generation and swimming speed via the caudal fin dynamics.

## I. Introduction

Fish species have evolved with astonishing success in diversification and locomotion capabilities in various underwater environments [1]-[5]. Fish swimming relies primarily on undulatory motions in body and fins and has provided design templates for novel underwater propulsion and vehicles aiming at achieving higher efficiency and maneuverability than those using conventional propellers [6]-[9].

The diversification of fish species renders complex forms and functions of swimming, from which it is challenging to extract general design templates and principles for robotic emulation. Fish swimming, by its propulsion mechanism, can be categorized into two types: Body and/or Caudal Fin (BCF)

Research supported by National Science Foundation (CNS-1932130 awarded to B.C) and Army Research Office (W911NF-20-1-0226, awarded to B.C).

[1]Department of Mechanical Engineering, Penn State University, University Park, PA, 16802, USA. dpl5368@psu.edu, buc10@psu.edu.

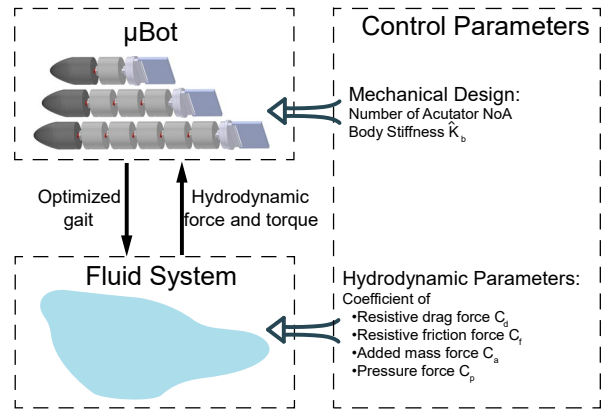

Fig. 1. Overview of this work. We explored the effect of control parameters in actuator number, body stiffness, and hydrodynamic parameters ($C_a$, $C_p$, $C_f$, $C_d$).

propulsion and Median and/or Paired Fin (MPF) propulsion [10]. While BCF forms achieve higher speed and efficiency in cruising, MPF forms offer better maneuverability [11]. As 85% of fish families are estimated to use BCF propulsion [3], it is not surprising that BCF has served as the major design template in fish-inspired robots. This template of fish swimming primarily composes of a smooth, elongated body with caudal fin attached posteriorly, while fins along the body such as pectoral fins and dorsal fins can be considered as secondary features. Even within this parsimonious design template, there exist many variations such as in caudal fin shape and stiffness, body slenderness, stiffness, mass distribution, and number of actuation (or effective links). These design variations, when coupled with various fluid mechanisms for hydrodynamic force generation via large possible swimming gaits, render the robot-fluid interaction problem difficult to solve or be included in the design process.

Fish or fish-inspired swimming are often studied using both experiments with biological [12], [13] or robotic fish [14], [15] and computational simulation [16]-[18]. Experiments with robotic fish can directly reveal the swimming performance with the actual physics of robot-fluid interaction. However, these experimental setups are costly in prototyping with large and frequent changes in mechanical design for systematic exploration in the design space. Computational Fluid Dynamic (CFD) simulation can solve Navier-Stokes equations based on known swimming kinematics and sometimes the entire fluid-structure interaction in simplified form. However, despite its high accuracy, CFD is computationally costly for systematic exploration of design space as well. Therefore, to systematically investigate how the variations within the fish-inspired design template affect the swimming gaits and performance, building an accurate and computationally efficient mathematical model for swimming is the most feasible approach.

The most widely used method for modeling fish swimming is based on the large amplitude elongated body theory (LAEBT) [19] or Lighthill fish swimming model. In the LAEBT, 3D problem is simplified as a 2D problem with potential flow and reactive force is calculated based on the fluid in a control volume attached to the fish. As compared to the CFD solution, LAEBT predictions often have acceptable accuracy with significantly lower computational cost. In recent decades, there are also several attempts to improve the Lighthill fish swimming model and adapt it to a mobile multi-link system [20].

In this work, we aimed to systematically explore the relationship among robot design properties, fluid dynamics, swimming gaits, and swimming performance in fish-inspired swimming. We first developed models and a simulation platform of Magnetic, Modular, Undulatory Robots (µBots) [21], which represents a robot template of fish-inspired swimming. We created total 36 simulation cases with distinct model parameters, namely number of actuators (NoA) and body stiffness, and hydrodynamic parameters. We optimized the swimming gaits for maximizing swimming speed and analysis the thrust generation mechanism. The rest of this paper is organized as follows. In section II, the design, mathematical modeling, simulation, and method of gait optimization of µBot are described. Section III presents the results from simulation. Discussions and future work are presented in section IV.

## II. MATERIALS AND METHODS

In this section, the design, mathematical modeling, simulation setup, and method of gait optimizations of µBot are described. In addition, the dimensional parameters that determine the physics of swimming are derived.

### A. Robot design and actuator model

The design details of µBot are introduced in our previous work [21], and only those that are relevant to modeling and simulation are provided here. The µBot is composed of a head segment, multiple body segments, a peduncle segment and a caudal fin (Fig. 2(a)). A body segment has an elliptical transverse plane with a nominal depth of 13.7 mm and width of 7 mm, and a rectangular sagittal plane with nominal aspect ratio of 2 (i.e., a nominal length of 27.4 mm). Caudal fin is mounted on the peduncle segment via a torsion spring and is modeled as a rigid plate with the same body depth and a thickness of 0.97mm. µBot is assumed to have a uniform density of 1.0kg/m³ with neutral buoyancy in the simulation. Except for the last posterior joint (peduncle-caudal fin), all other anterior joints are actuated by a magnetic actuator with a parallel torsion spring.

Each magnetic actuator (Fig. 2(b)) has a coil mounted on a rotating arm (coil clamp) around a pivot joint, while the coil is placed in between two permanent magnets pointed closely to each other with identical polarity (therefore opposing each other). With voltage applied, the coil generates lateral forces, which in turn create actuation torque via the rotating arm that rotates the subsequent segment and reverse the voltage simply reverses the torque.

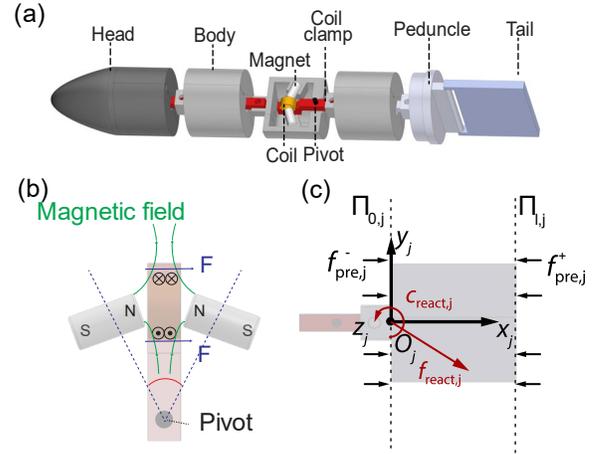

Fig. 2. Design and modeling of µBot. (a) Top view of µBot with 4 actuators. The interior actuator design of the 3rd segment (actuator) is shown. (b) Details of actuator design showing the permanent magnets and coil. (c) Added-Mass Volume (AMV) and reactive forces of one segment. $s$ and $n$ are segmental normal and longitudinal unit vectors.

To simulate the actuator, a mathematical model of which was developed by calculating the magnetic field of the permanent magnets[23] and the coil. This model is calibrated by the torque measured experimentally at zero displacement. The model showed that both the actuator torque constant ($k_\text{T}$) and back EMF constant ($k_\text{EMF}$) remained nearly constant across the usable range of coil displacement or the actuator angle. Therefore, actuator was modeled simply as,

$$T = \frac{E - k_\text{EMF} \cdot \omega_\text{joint}}{R} \cdot k_\text{T}, \qquad (1)$$

where $T$ is torque generated by actuator, $E$ is applied voltage, $\omega_{joint}$ is joint's angular velocity, $R$ is the resistivity of coil, torque constant is chosen as 1.26 Nmm/A, and back EMF constant is chosen as 1.26 mV · s.

### B. Modeling of hydrodynamic forces

To simulate the swimming of µBot, a model of hydrodynamic force was developed based on reactive theory[19], and resistive theory [22], following the approach used by Mathieu et al. [20] with minor revisions. Since µBot was neutrally buoyant, only horizontal planar motion and relevant forces and torque were considered.

We attached a mobile frame $(O_j, x_j, y_j, z_j)$ to the segment $j$. The unit vector $x_j$ is along segment longitudinal direction, $y_j$ is along segment lateral direction, and $z_j$ is along vertical direction. For any physical variable modeled as vectors, with j in lower index present the body index (to which it is related) and indicate the index of projection frame.

### B1. Reactive force model

Applying Lighthill's theory with potential flow assumption, we first defined a control volume attached to each segment, termed as added-mass volume (AMV). Each AMV has two boundaries $\Pi_0$ and $\Pi_l$, which are infinite planes fixed at the anterior and posterior side of each segment, respectively (Fig. 2(c)). The total segmental reactive force was modeled with two components: 1) added-mass force ($f_{add}$) due to the rate of change of fluid momentum within AMV, and 2) the pressure force -$f_{pre}^-$, $f_{pre}^+$ acting on the boundaries of AMV. For a given robot segment $j$ (Fig. 2c)), the total reactive wrench $F_{react,j}$, exerted by the fluid within AMV, can be written as

$$F_{react,j} = (f_{react,j}^T, c_{react,j}^T)^T. \quad (2)$$

Here $F_{react,j}$ in $R^6$ is wrench and includes both force ($f_{react,j}^T$) and torque ($c_{react,j}^T$) components (about each joint). Applying Newton's laws and Euler's theory to the fluids in AMV, it can be shown that,

$$f_{react,j} = \frac{d}{dt}\iiint_{AMV} P_j \cdot dV_{AMV} + f_{pre,j}^- - f_{pre,j}^+, \quad (3)$$

$$c_{react,j} = -\frac{d}{dt}\iiint_{AMV} \Sigma_j \cdot dV_{AMV}, \quad (4)$$

where $P_j$ and $\Sigma_j$ are fluid's linear momentum and angular momentum of one slice of fluid in AMV. In Lighthill's original theory, fluid outside AMV was assumed to have no velocity, so that $f_{pre,j}$ can be calculated using unsteady Bernoulli's principle. However, this assumption was biased as fluid outside boundaries still has velocity and pressure force was over-estimated. In this work, we introduced a correction coefficient $C_p$ for pressure force. Evaluating (3) and (4), segmental reactive force becomes,

$$F_{react,j} = \begin{bmatrix} -\bar{m}_j l_j \omega_j v_{0,j} + \frac{1}{2}\bar{m}_j l_j^2 \omega_j^2 + C_p \cdot \frac{1}{2}\bar{m}_j(v_{0,j}^2 - v_{l,j}^2) \\ -\bar{m}_j l_j a_{0,j} - \frac{1}{2}\bar{m}_j l_j^2 \dot{\omega}_j \\ 0 \\ 0 \\ 0 \\ -[\frac{1}{2}\bar{m}_j l_j^2(a_{0,j} - \omega_j u_j) + \frac{1}{3}\bar{m}_j l_j^3 \dot{\omega}_j + \bar{m}_j l_j u_j v_{l,j}] \end{bmatrix} \quad (5)$$

where $\bar{m}_j$ is the cross-sectional added mass, $l_j$ is segment length, $\omega_j$ is segment's angular velocity, $u_j$ is segment's longitudinal velocity, $v_{0,j}$ is segment's lateral velocity at boundary $\Pi_{0,j}$, $a_{0,j}$ is segment's lateral acceleration at boundary $\Pi_{0,j}$, and $v_l$ is segment's lateral velocity at boundary $\Pi_{l,j}$. According to Lighthill [24], $\bar{m}$ is defined as

$$\bar{m}_j = C_a \cdot \frac{1}{4}\pi h_j^2 \rho_f \quad (6)$$

where $h_j$ is segment depth, $\rho_f$ is fluid density, and $C_a$ is coefficient of added mass.

### B2. Resistive force model

Based on Taylor's theory [22], resistive force $F_{resis}$ takes the form of lateral (normal) drag force and longitudinal (tangential) friction force,

$$F_{resis,j} = -\frac{1}{2}\rho_f \begin{bmatrix} C_f \cdot P_j \cdot l_j \cdot |u_j|u_j \\ C_d h_j \int_0^{l_j}|v_{x,j}|v_{x,j}dx_j \\ 0 \\ 0 \\ 0 \\ C_d h_j \int_0^{l}|v_{x,j}|v_{x,j} \cdot x dx_j \end{bmatrix}, \quad (7)$$

where $P_j$ is segment perimeter, $C_f, C_d$ are friction and drag coefficients that are Reynolds-number dependent, $u_j$ is segment's longitudinal velocity and $v_{x,j}$ is lateral velocity of slice of fluid at longitudinal coordinate $x$.

### C. Dimensionless parameters (control parameters) of swimming and robot-fluid interaction

Fish swimming involves complex interactions between the fish's body and the surrounding fluids, posing an inherent fluid-structure interaction (FSI) problem. Here, we derive a number of dimensionless parameters that effectively "control" the nature of this interaction in addition to the gait parameters (which are optimized). Dimensionless variables (unit $\hat{x}$, time $\hat{t}$, longitudinal velocity $\hat{u}$, lateral velocity $\hat{v}$, angular velocity $\hat{\dot{\theta}}$ and angular acceleration $\hat{\ddot{\theta}}$) were defined as,

$$\hat{x}_j = \frac{x}{l_j}, \; \hat{t} = ft, \; \hat{u} = \frac{u}{U}, \; \hat{v} = \frac{v}{f\bar{A}}, \; \hat{\dot{\theta}} = \frac{\dot{\theta}}{f}, \; \hat{\ddot{\theta}} = \frac{\ddot{\theta}}{f^2}, \quad (8)$$

where $f$ is undulatory frequency, $U$ is average cruising speed, $\bar{A}$ is segment's average lateral displacement. The dimensionless model of hydrodynamic force are,

$$\begin{bmatrix} \frac{f_{react}}{\frac{\pi}{4}\rho_f l^2 h^2 f^2} \\ \frac{c_{react}}{\frac{\pi}{4}\rho_f l^3 h^2 f^2} \end{bmatrix} = C_a \cdot \begin{bmatrix} \frac{\bar{A}}{l}\hat{\dot{\theta}}\hat{v} + \frac{1}{2}\hat{\dot{\theta}}^2 + C_p\left(\frac{\bar{A}}{l}\hat{v}\hat{\dot{\theta}} - \frac{1}{2}\hat{\dot{\theta}}^2\right) \\ -\frac{\bar{A}}{l}\hat{a} - \frac{1}{2}\hat{\ddot{\theta}} \\ 0 \\ 0 \\ 0 \\ \frac{1}{2}\frac{\bar{A}}{l}\hat{a} - \frac{1}{2}\frac{U}{lf}\hat{u}\hat{\dot{\theta}} + \frac{1}{3}\hat{\ddot{\theta}} + \frac{U}{lf}\frac{\bar{A}}{l}\hat{u}\hat{v} \end{bmatrix}, \quad (9)$$

$$\begin{bmatrix} \frac{f_{resis}}{\frac{\pi}{4}\rho_f l^2 h^2 f^2} \\ \frac{c_{resis}}{\frac{\pi}{4}\rho_f l^3 h^2 f^2} \end{bmatrix} = -\frac{2}{\pi} \cdot \begin{bmatrix} C_f \cdot AR \cdot \left(\frac{U}{lf}\right)^2 |\hat{u}|\hat{u} \\ C_d \cdot AR \cdot \left(\frac{\bar{A}}{l}\right)^2 \int_0^1 |\hat{V}|\hat{V}d\hat{s} \\ 0 \\ 0 \\ 0 \\ C_d \cdot AR \cdot \left(\frac{\bar{A}}{l}\right)^2 \int_0^1 |\hat{V}|\hat{V}\hat{x}d\hat{x} \end{bmatrix}, \quad (10)$$

where $\hat{V} = \hat{v} + \frac{l}{\bar{A}}\hat{x}\hat{\dot{\theta}}$ and $AR = \frac{l}{h}$ is body segmental aspect ratio.

From the dimensionless model, we identified two types of dimensionless parameters: 1) those that need to be defined *a priori* before starting simulation and optimization, such as $AR, C_a, C_p, C_f, C_d$, which are the 'control parameters' of the swimming physics, and 2) those that will be obtained after gait optimization, such as $\frac{\bar{A}}{l}, \frac{U}{lf}$.

To investigate the effects of hydrodynamic parameters, which change with the Reynolds number, we created four hydrodynamic models (HM), with different combinations of $C_p$ and $C_a$ (Table 1). HM-1 represents a model in which robot only experiences resistive force from fluid due to boundary layer effects (or circulatory-based frictional and drag forces). HM-2 represents a model containing resistive force and part of the reactive force with $C_p = 0$. In this model, we assume there is no difference in fluid velocity between both sides of boundaries of AMV, therefore pressure force -$f_{pre}^-$ and $f_{pre}^+$ are zero. HM-3 denotes a model containing all resistive force and reactive force components with $C_p = 0.5$. HM-4 denotes a classical hydro-dynamic model that contains all resistive force and reactive force components with $C_p = 1$. From HM-2 to HM-4, we gradually increase the coefficient of pressure force to explore how it affects robot swimming.

TABLE I. SETTING OF CONTROL PARAMETERS

| Control Parameters | Values of Control Parameters | | | |
|---|---|---|---|---|
| Number of actuators NoA | 2 | 4 | | 6 |
| Normalized Stiffness $\widehat{K}_b$, (Nmm/rad) | High $\frac{1.00}{16}$ | Medium $\frac{0.75}{16}$ | | Low $\frac{0.50}{16}$ |
| Hydrodynamic Model | HM-1 $C_a=0$ $C_p=0$ | HM-2 $C_a=1$ $C_p=0$ | HM-3 $C_a=1$ $C_p=0.5$ | HM-4 $C_a=1$ $C_p=1$ |

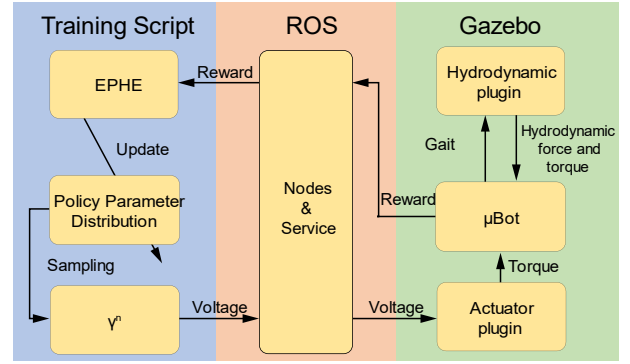

Fig. 3. Schematic view of the simulation platform.

The effects of two body design parameters were investigated: 1) number of actuation (NoA) or the number of body segments, as the μBot is a multilink robot, and 2) spring stiffness ($K_b$). For an N-actuator μBot model, $K_b$ can be written as,

$$K_b^j = \widetilde{K}_b \cdot \Gamma(j), \qquad j = 1,2 \dots N+1, \qquad (11)$$

where $K_b^j$ is torsional spring stiffness for $j^{th}$ joint and $\Gamma(j)$ is stiffness distribution function. Here we assume,

$$\Gamma(j) = \begin{cases} 1, & j \neq N+1 \\ 5, & j = N+1 \end{cases}, \qquad (12)$$

i.e., the caudal fin joint stiffness is 5 times of body stiffness. To isolate the effects of $\widetilde{K}_b$ from $AR$, we defined the normalized spring stiffness $\widehat{K}_b$,

$$\widehat{K}_b = \frac{\widetilde{K}_b}{AR^4}. \qquad (13)$$

In total, $AR$, $(C_f, C_d)$, $HM$, $\widehat{K}_b$ and NoA are five key dimensionless parameters, or Control Parameters (CPs). In this study, we investigated the effects these CPs (except the $\widehat{K}_b$) by creating total 36 simulations cases according to Table 1.

*D. Gait Optimization*

As the CPs for mechanical design and hydrodynamic model are set, the next problem is gait optimization. Instead of directly using the swimming gait from biological fishes, here we optimized μBot's gaits for swimming speed by optimizing actuator voltage signals. In this work, we generated the voltage signal $E_j(t)$ sent to $j^{th}$ actuator according to,

$$E_j(t) = \begin{cases} e_j \sin(2\pi f_{input} t), & j = 1 \dots \\ e_j \sin(2\pi (f_{input} t + \Psi_j)), & j = 2 \dots NoA \end{cases} \qquad (14)$$

where $e_j$ is voltage amplitude, $\Psi_j$ is initial phase of voltage, and $f_{iput}$ is input signal frequency. Given this definition, the optimization parameter vector $\gamma$, that governing voltage signal, can be constructed as,

$$\gamma = [e_1, \Psi_2, e_2 \dots \Psi_{NoA}, e_{NoA}, f]. \qquad (15)$$

In this work, a policy gradient-based reinforcement learning (RL) method, i.e., EM-based Policy Hyper Parameter Exploration (EPHE) [25] was used to optimize the swimming gaits of swimming. In EPHE, policy parameters $\gamma$ is sampled from probability distribution $p(\gamma|\eta, \sigma^2)$ where $\eta, \sigma$ are the hyperparameter vectors composed of mean value and standard deviation of probability distribution. To obtain a good sampling performance, only the policy parameters from the best $K$ rollouts (trajectories with highest $K$ rewards) in total $M$ rollouts are taken for updates. Hyperparameters are updated as,

$$\eta = \frac{\sum_{i=1}^{K}[R(\gamma^i)\gamma^i]}{\sum_{i=1}^{K} R(\gamma^i)}, \qquad (16)$$

$$\sigma = \sqrt{\frac{\sum_{i=1}^{K}[R(\gamma^i)(\gamma^i-\eta)^2]}{\sum_{i=1}^{K} R(\gamma^i)}} \qquad (17)$$

where $R(\gamma^i)$ is the reward of the $i^{th}$ rollout.

Empirically, μBot took less than 2 seconds for acceleration. Therefore, we set μBot to swimming for 6 seconds and used the average speed within the last 2 seconds as the reward of learning.

*E. Robot Simulation Setup*

Next, the above hydrodynamic models and EPHE algorithm were used to create a simulation platform, which was composed of three parts: Gazebo, training script, and

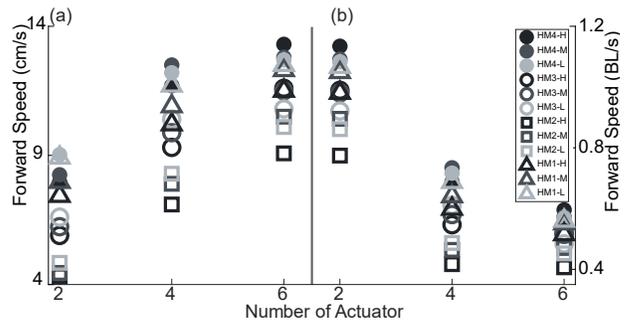

Fig. 4. Optimized forward speed of all 36 cases. 'H', 'M' and 'L' denote high stiffness, medium stiffness and low stiffness. (a) shows the absolute optimized forward speed for all 36 cases. (b) shows the optimized forward speed in body-length per second (BL/s).

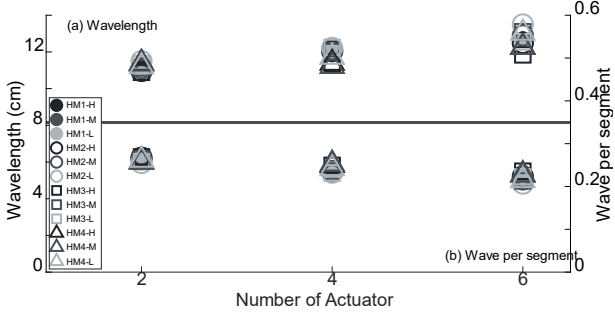

Fig. 5. Wavelength of optimized gaits remained approximately constant from 36 cases. (a) shows optimized wavelengths for all 36 cases with scales on left. (b) shows optimized wave per segment for all 36 cases with scales on right.

Robot Operating System (ROS). Gazebo, employing plugins with hydrodynamic model and actuator model, handles the simulation of robot dynamic, calculation of hydrodynamic force and generation actuators' torque. Training script contains learning algorithm and generates signals for actuators. To integrate the training script into the Gazebo simulator, we used ROS for action controlling and data passing. During the simulation, training script uses EPHE to sample policy parameters, generates voltage signals and sends signals to Gazebo via ROS nodes and services. Then, Gazebo calculates robot dynamics and record robot kinematic information. Once one simulation rollout is done, Gazebo sends reward to training scrip via ROS nodes and services (Fig. 3).

## III. RESULTS

In the current work, we investigated the effects of NoA, hydrodynamic coefficients of reactive force (four combinations of $C_a$ and $C_p$) and stiffness $\widehat{K}_b$. The remaining control parameters were fixed $AR=2$, $C_f=0.06$, and $C_d=2.25$. To train μBot model, in each episode we performed 50 rollouts and picked the 25 best performing rollouts to update the actuation policy. In each training session we have updated the policy 40 times. For each simulation case we repeated the training sessions 3 times with random initial conditions. The repeated training sessions converged similar final reward values (forward speeds) with only negligible differences.

### A. Effects of number of actuators (NoA)

The (optimized) absolute forward speed increases with NoA (Fig. 4a), however the forward speed per body-length decreases with NoA, suggesting diminishing gain of forward speed from adding more actuators or body segments. This trend is also independent of the hydrodynamic models (with different combinations ($C_a$, $C_p$)) or spring stiffness of $\widehat{K}_b$.

Fig. 5 shows the dependence of optimized gait wavelength on NoA. The wavelengths remain approximately constant at 12 cm regardless of the robot length and the averages of wave per body segment remain approximately 0.25, i.e., a full wave spans 4 body segments. Unlike observations in biological fish, which shows a positive correlation between wavelength and body length [26], our results indicate that wavelength of optimized swimming in μBots is only weakly correlated to the selected control parameters, NoA, hydrodynamic parameters and body stiffness.

### B. Effects of hydrodynamic parameters

Fig. 6 shows the total body thrust generation (summing across body segments and caudal fin) and tail torque generation due to different hydrodynamic mechanisms for all cases. From Fig. 6(a) to Fig. 6(c), resistive friction force is the dominant force resisting the motion along the swimming direction. In steady state swimming, the resistive friction force is balanced by resistive drag force in HM-1, by resistive drag force (primary) and added-mass force (secondary) in HM-2, by caudal-fin pressure forces (primary) and resistive drag and added-mass forces (both secondary) in HM-3 and HM-4. Note that the pressure force, being the dominant mechanism for thrust generation, is mainly from the caudal

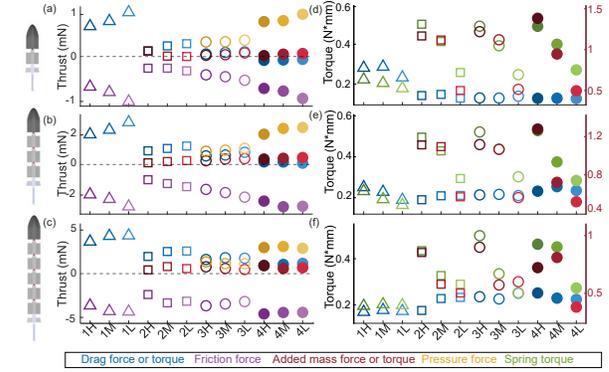

Fig. 6. Total body thrust generation due to different hydrodynamic mechanisms and torques applied at tail in all cases. On x axis, '1', '2', '3', '4' denote HM-1, HM-2, HM-3, HM-4 and 'H', 'M', 'L' denotes high stiffness, medium stiffness and low stiffness. (a)(b)(c) shows total thrust generated from different type of force with 2 actuators, 4 actuators and 6 actuators. (d)(e)(f) shows torque applied tail with 2 actuators, 4 actuators and 4 actuators.

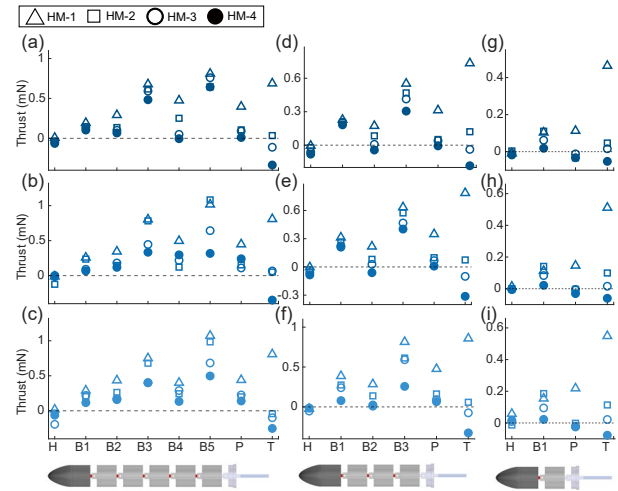

Fig. 7. Distribution of thrust from resistive drag force along μBot for all 36 cases. Four different hydrodynamic models are distinguished by shapes: triangles for HM-1 cases, squares for HM-2 cases, hollow circles for HM-3 cases and filled circles for HM-4 cases. Columns from left to right denotes thrust distribution with 6 actuators, 4 actuators, and 2 actuators. Rows from top to bottom shows thrust distribution with high stiffness, medium stiffness and low stiffness. In each plot, x axis denotes different segments, while 'H','B', 'P' and 'T' denote head segment, body segment, peduncle segment and tail segment,

fin (results not shown), indicating the importance of caudal-fin design and dynamics in thrust generation and swimming speed [27].

Fig. 7(a)-(i) further shows the distribution of thrust from resistive drag force along the body segments of μBots. Unlike the pressure force, the drag force from body segments contribute to the thrust generation, while the caudal-fin drag force has negligible contribution to thrust in all cases except for the case with NoA =2 (resembles thunniform swimmer) and HM-1.

While added-mass forces do not contribute significantly to thrust generation, the added-mass torque, on the other hand, dominates the caudal fin dynamics (Fig. 6(d) - Fig. 6 (f)). It can be seen that added-mass torque has similar magnitude to spring torque, indicating a balance in a near resonant state (except the low spring stiffness cases). Video S1 shows that μBot, under hydrodynamic model with (HM-2,HM-3,HM-4) and without (HM-1) added-mass forces, has categorically different caudal-fin kinematics, which lead to different thrust generation mechanisms. Therefore, added-mass force indirectly affects the thrust generation and swimming speed via the caudal fin dynamics.

In addition, our results show that the actuation torque, fluid added-mass torque and spring torque have similar magnitude on body segments including head and peduncle segments (results not shown. In contrast, the contributions from the resistive drag force was again negligible. This indicates that the robot gait generation, including both body and caudal-fin gaits, are dominated by the interaction between the robot dynamics and the reactive hydrodynamic forces.

IV. DISCUSSIONS AND FUTURE WORK

In summary, our results show that increasing NoA (i.e., increasing robot length and number of actuators) increases the optimized forward swimming speed, however, with a diminishing gain, as the forward speed per body length decreases. Note that among the 36 cases, the μBot with NoA = 6, HM-4 model and highest stiffness has the fastest forward speed of 13.3cm/s and 0.60BL/s and the Bot with NoA = 2, HM-2 model and high stiffness has the lowest forward speed of 4.32cm/s and 0.37BL/s.

Interestingly, via examining the contributions of different hydrodynamic mechanisms, we found fundamentally different roles of fluid added-mass towards thrust generation and gait generation: the added-mass *force* had negligible contribution to the thrust generation, while added-mass *torque* dominates the gait generation in both caudal-fin and body segments, together with spring torque and actuator torque. The resistive drag force, on the other hand, has negligible contribution towards the gait generation, however, contribute to the thrust mainly from the body segments. These understandings will be crucial for understanding the experimental data of μBot, including performing [21] robot system identification and possible model-based reinforcement learning in the future.

Finally, note that the current analysis does not fully explore the specific effects of body and caudal fin stiffness $\widehat{K}_b$, because only three sample values were used. Although the optimized swimming speed seem to only depend weakly on the stiffness, the swimming gait and efficiency could have greater dependence. In future work, we plan to perform a more comprehensive study on the effects of body and caudal-fin spring stiffness. Furthermore, in this work, we only examined the effects of three control parameters out of the five derived (see in II.C). In future work, we will continue to explore the effects of these control parameters and investigate the relationship among body morphologies, fluid dynamics, swimming gaits and swimming performance in fish-inspired swimming.